\documentclass{article}

    \PassOptionsToPackage{numbers, compress}{natbib}

    \usepackage[preprint]{neurips_2023}

\usepackage[utf8]{inputenc} 
\usepackage[T1]{fontenc}    
\usepackage{hyperref}       
\usepackage{url}            
\usepackage{booktabs}       
\usepackage{amsfonts}       
\usepackage{nicefrac}       
\usepackage{microtype}      
\usepackage{xcolor}         

\usepackage{amsmath}
\usepackage{graphicx}
\usepackage{caption}
\usepackage{subcaption}

\newcommand{\draftonly}[1]{#1}
\renewcommand{\draftonly}[1]{}

\newcommand\Tstrut{\rule{0pt}{2.6ex}}         
\newcommand\Bstrut{\rule[-0.9ex]{0pt}{0pt}}   

\title{Probing Intersectional Biases in Vision-Language Models with Counterfactual Examples}

\author{%
  Phillip Howard \\
  Intel Labs \\
  \texttt{phillip.r.howard@intel.com} \\
  \And
  Avinash Madasu \\
  Intel Labs \\
  \texttt{avinash.madasu@intel.com} \\
  \And
  Tiep Le \\
  Intel Labs \\
  \texttt{tiep.le@intel.com} \\
  \And
  Gustavo Lujan Moreno\\
  Intel Labs \\
  \texttt{gustavo.lujan.moreno@intel.com} \\
  \And
  Vasudev Lal \\
  Intel Labs \\
  \texttt{vasudev.lal@intel.com} \\
}

\begin{document}

\maketitle

\begin{abstract}

While vision-language models (VLMs) have achieved remarkable performance improvements recently, there is growing evidence that these models also posses harmful biases with respect to social attributes such as gender and race. Prior studies have primarily focused on probing such bias attributes individually while ignoring biases associated with intersections between social attributes. This could be due to the difficulty of collecting an exhaustive set of image-text pairs for various combinations of social attributes from existing datasets. To address this challenge, we employ text-to-image diffusion models to produce counterfactual examples for probing intserctional social biases at scale. Our approach utilizes Stable Diffusion with cross attention control to produce sets of counterfactual image-text pairs that are highly similar in their depiction of a subject (e.g., a given occupation) while differing only in their depiction of intersectional social attributes (e.g., race \& gender). We conduct extensive experiments using our generated dataset which reveal the intersectional social biases present in state-of-the-art VLMs. 

\end{abstract}

\section{Introduction}

Counterfactual examples, which study the impact on a response variable following a change to a causal feature, have proven valuable in natural language processing (NLP) for probing model biases and robustness to spurious correlations. While counterfactual examples for VLMs have been relatively unexplored, recent work has shown that text-to-image diffusion models with cross attention control can effectively produce multimodal counterfactual examples for VLM training data augmentation and evaluation \citep{le2023coco}. This suggests that synthetic counterfactual examples generated by diffusion models could be an effective tool for probing bias in VLMs.

Bias in pre-trained models can be viewed as spurious correlations attributed to the co-occurrence of non-causal features with target labels in datasets. During pre-training, models learn to exploit such correlations as shortcuts to achieving high in-domain performance on the training dataset \citep{geirhos2020shortcut}. Consequently, models which learn to rely on spurious correlations are more brittle and have worse out-of-domain (OOD) generalization \citep{singla2021salient, xiao2020noise}. 

Social biases are a particularly concerning type of spurious correlation learned by VLMs. Due to a lack of proportional representation for people of various races, genders, and other social attributes in image-text datasets \citep{birhane2021multimodal}, VLMs learn biased associations between these attributes and various subjects (e.g., occupations). For example, given a gender- and race-neutral query such as ``A photo of an attorney'', a VLM may retrieve a disproportionate number of images of white men due to learned spurious correlations between this particular occupation and the intersection of race-gender attributes. 

Prior studies on probing social biases in VLMs have primarily utilized real image-text pairs collected from existing datasets by identifying the co-occurrence of certain attributes with a target subject. However, this approach is limited by the availability of existing image-text pairs for various combinations of social attributes and subject types. Previous work has consequently focused exclusively on investigating biases associated with a single social attribute at a time while ignoring the potential role of intersectional bias (e.g., particular race-gender combinations) \citep{navigli2023biases}, which could be attributed to the difficultly of collecting an exhaustive set of image-text examples for various combinations of social attributes. Additionally, the large variability in how subjects can be naturally depicted in real images complicates the task of estimating bias in VLMs because disproportionate retrieval results could potentially be attributed to other differences in images besides the social attribute. 

\begin{figure}[t!]
    \centering
    \includegraphics[trim={2mm 2mm 2mm 
    2mm},clip,width=1\columnwidth]{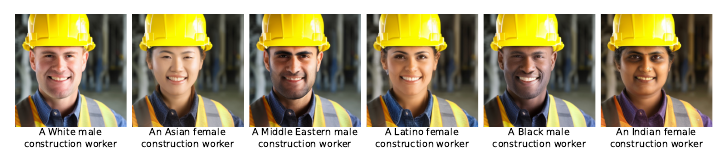}
    \caption{Examples of our counterfactual image-text pairs for probing intersectional race-gender bias in VLMs for the ``construction worker'' occupation. See Appendix~\ref{app:additional-examples} for additional examples.
    }
    \label{fig:main_examples}
\end{figure}

We overcome these limitations by leveraging text-to-image diffusion models to produce counterfactual image-text pairs for probing social biases in VLMs (see Figure~\ref{fig:main_examples} and Appendix~\ref{app:additional-examples} for examples). Specifically, our approach utilizes Stable Diffusion \citep{rombach2021highresolution} with cross-attention control \citep{hertz2022prompt} to produce a set of highly similar counterfactual image-text examples which depict a common subject while differing only in intersectional social attributes. 
We apply our methodology at scale to produce an extensive dataset for evaluating intersectional social biases and conduct experiments to uncover such biases in state-of-the-art VLMs.\footnote{We will make our dataset and code publicly available to support future research}

\section{Methodology}

Our approach to creating counterfactual image-text examples for probing social biases consists of two steps. First, we construct sets of image captions describing a subject with counterfactual changes to intersecting social attributes. We then utilize a text-to-image diffusion model with cross attention control to generate a set of images corresponding to the counterfactual captions, where differences among images are isolated to the induced counterfactual change (i.e., the social attributes).

\subsection{Constructing captions for probing social biases}

Consider the task of creating a caption $C^{s}_{p, a_{1},a_{2}}$ beginning with prefix $p$ and describing a subject $s$ which possesses a pair of attributes $a_{1}$ and $a_{2}$. Given a set of prefixes $P$, a set of subjects $S$, and attribute sets $A_{1}, A_{2}, ..., A_{k}$, we populate the following template to obtain our captions:
\begin{equation*}
\begin{aligned}
    C^{s}_{p, a_{1},a_{2}} = \textrm{<$p$> <$a_{1}$> <$a_{2}$> <$s$>} && \forall \ p \in P, s \in S, a_{1} \in A_{i}, a_{2} \in A_{j}, (i, j) \in \{1, ..., k \ | \ i \ne j\} \\
\end{aligned}
\end{equation*}

We construct captions in this manner using two subject sets (occupations and personality traits) and four sets of attributes for measuring social bias (gender, race, religion, and physical characteristics). 
Captions are grouped into counterfactual sets, where each set contains all captions corresponding to a given prefix and subject. Using 261 occupations, 63 personality traits, 6 races, 4 religions, 2 genders, and 14 physical characteristics, we produced a total of 232k captions which were grouped into 7k counterfactual sets (see Appendix~\ref{app:caption-details} for additional details and examples). 

\subsection{Counterfactual image generation}

After generating sets of counterfactual captions, we use text-to-image diffusion models to produce images for each caption. In order to precisely measure the impact of social attribute differences, it is desirable for images within a counterfactual set to only differ in how the social attributes differ across captions. However, this is challenging for diffusion models as even minor changes to a prompt can result in the generation of images with significant differences. For example, changing the attributes \textit{Hispanic female} to \textit{Asian male} in the prompt \textit{A photo of a Hispanic female doctor} may produce other undesired modifications to the image that extend beyond the induced counterfactual change (e.g., changes to the background). This complicates the task of quantifying the impact of model bias attributed to the changed social attributes on retrieval results, as other differences between the generated images could contribute to a VLM's preference for retrieving particular images.

\citet{hertz2022prompt} proposed Prompt-to-Prompt to address this issue by injecting cross-attention maps during denoising steps to control attention between certain pixels and tokens, which enables separate generations to maintain many of the same details while isolating differences to how the text prompts differ. 
However, \citet{brooks2023instructpix2pix} noted that some changes require varying the parameter $p$ in Prompt-to-Prompt, which controls the number of denoising steps with shared attention weights. 
We therefore adopt their proposed approach of over-generating 100 image pairs with Prompt-to-Prompt by sampling $p \sim U(0.1,0.9)$. The 100 image pairs are filtered using CLIP \citep{radford2021learning} to ensure a minimum cosine similarity of 0.2 between the encoding of each caption and its corresponding generated image, with the best image pair chosen according to the directional similarity in CLIP space \citep{gal2022stylegan}

We extend Prompt-to-Prompt for image pairs from \citet{brooks2023instructpix2pix} to support batched generation of multiple images with shared cross cross-attention maps. This enables simultaneous generation of entire sets of counterfactual images which differ only according to the social attribute differences across prompts. In total, we generate 23.2M images for 232k captions. After filtering with CLIP, we keep up to 10 of the highest-scoring counterfactual sets according to CLIP directional similarity.

\section{Evaluation Metrics}

\begin{figure}[t!]
    \centering
    \begin{subfigure}[b]{0.49\textwidth}
    \includegraphics[trim={2mm 0mm 2mm 
    2mm},clip,width=1\columnwidth]{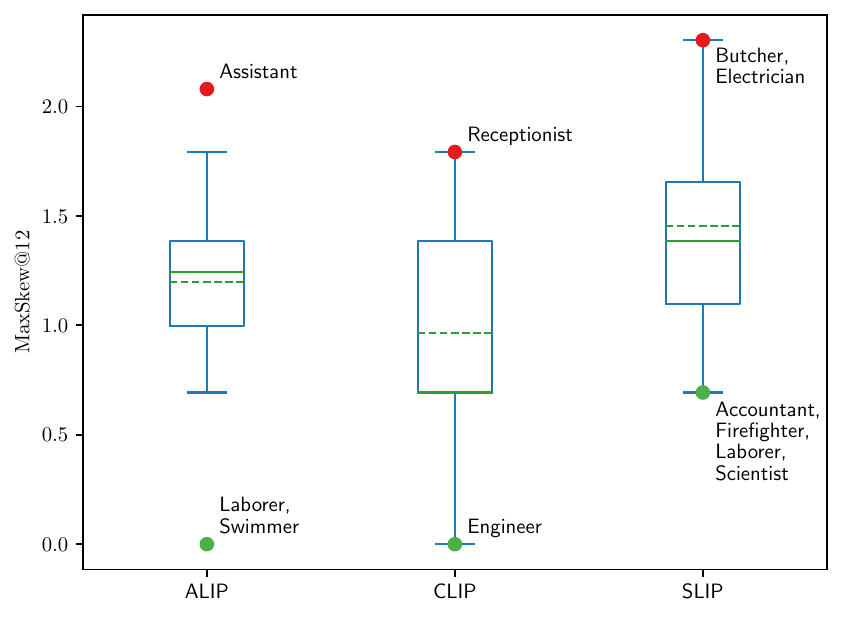}
    \end{subfigure}
    \begin{subfigure}[b]{0.475\textwidth}
    \includegraphics[trim={2mm 2mm 2mm 
    2mm},clip,width=1\columnwidth]{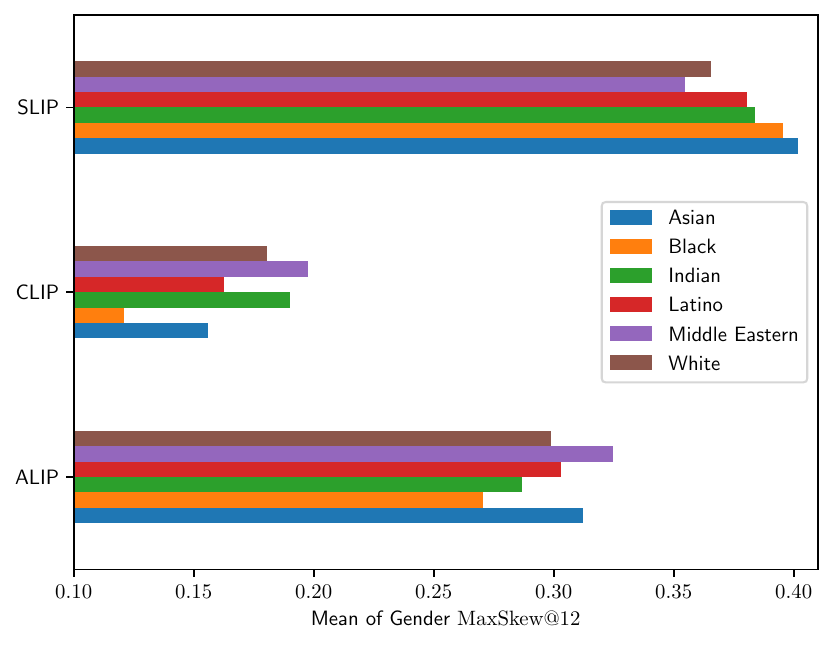}
    \end{subfigure}
    \caption{
    Left: Distribution of $\textrm{MaxSkew}@12$ measured across occupations for intersectional race-gender bias. Max (min) values are plotted as red (green) circles with corresponding occupation names. Right: Mean of (marginal) gender $\textrm{MaxSkew}@12$ measured across occupations for different races. 
    }
    \label{fig:maxskew}
\end{figure}


\label{sec:metrics}

Let $q$ denote a text query and $R_{K}(q)$ denote the set of top-$K$ ranked images retrieved by a VLM for $q$. For a given attribute pair $(a_{i}, a_{j})$, we denote the desired proportion of retrieved images with the corresponding attributes as $p_{d(q), (a_{i}, a_{j})}$ and the actual proportion as $p_{R_{K}(q), (a_{i}, a_{j})}$. \citet{geyik2019fairness} define $\textrm{Skew}@K$ for attributes $(a_{i}, a_{j})$ in retrieval results $R_{K}(q)$ as:
\begin{equation}
    \textrm{Skew}_{(a_{i},a_{j})}@K(R_{K}(q)) = \log(\frac{p_{R_{K}(q), (a_{i}, a_{j})}}{p_{d(q), (a_{i}, a_{j})}})
\end{equation}
In essence, $\textrm{Skew}@K$ measures the ratio of the proportion of top-$K$ retrieved images having a set of attributes to the desired proportion. To aggregate $\textrm{Skew}@K$ over the various attributes under consideration, \citet{geyik2019fairness} further proposed the following $\textrm{MaxSkew}@K$ metric:
\begin{equation}
    \textrm{MaxSkew}@K(R_{K}(q)) = \max_{(a_{i}, a_{j}) \in A} \textrm{Skew}_{(a_{i},a_{j})}@K(R_{K}(q))
\end{equation}
where $A$ denotes the set of all attribute pairs. We calculate $\textrm{MaxSkew}@K$ by retrieving images from our counterfactual sets using prompts which are neutral with respect to the investigated attributes. For example, given a prompt constructed from the template ``A <race> <gender> construction worker'' (Figure~\ref{fig:main_examples}), we form its corresponding attribute-netural prompt ``A construction worker.'' 

We construct neutral prompts in this manner for each unique combination of prefixes and subjects, averaging their CLIP representations across different prefixes to obtain a single text embedding for each subject. $\textrm{Skew}@K$ and $\textrm{MaxSkew}@K$ are then calculated by retrieving the top-$K$ images for the computed text embedding from the set of all images generated for the subject which met our filtering and selection criteria. We set $K = |A_{1}| \times |A_{2}|$, where $A_{1}$ and $A_{2}$ are the investigated attribute sets.

\section{Results}
\label{sec:results}

\begin{figure}[t!]
    \centering
    \includegraphics[trim={2mm 2mm 2mm 
    2mm},clip,width=1\columnwidth]{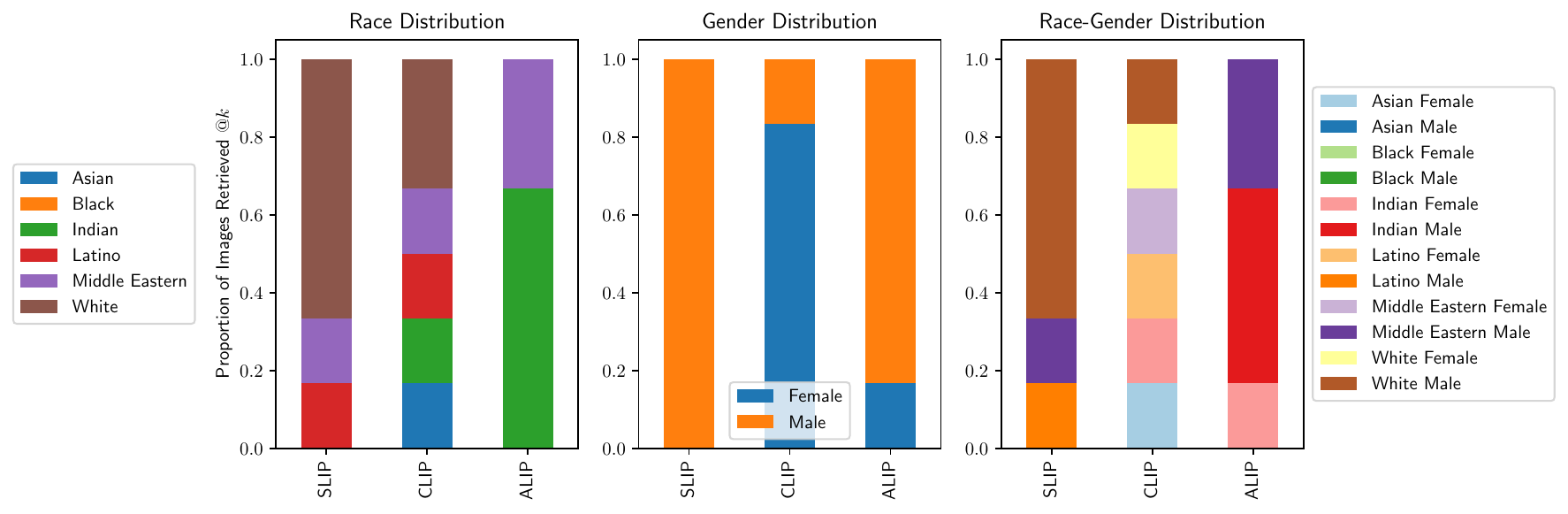}
    \caption{
    Proportion of images retrieved $@k=12$ using neutral prompts for the `Doctor' occupation.
    }
    \label{fig:doctor}
\end{figure}

Figure~\ref{fig:maxskew} (left) provides boxplots of the $\textrm{MaxSkew}@K$ distribution measured across occupations for intersectional race-gender bias attributes using ALIP \citep{yang2023alip}, CLIP \citep{radford2021learning}, 
and SLIP \citep{mu2022slip}.
All three VLMs exhibit skewness in retrieval for gender- and race-neutral occupation prompts, with CLIP having the lowest mean $\textrm{MaxSkew}@K$ and SLIP having the highest. 
Notably, ALIP exhibited zero skewness for only two occupations (`Laborer' and `Swimmer'), CLIP exhibited zero skewness for only one occupation (`Engineer'), and SLIP exhibited positive skewness across all occupations.

We estimated the marginal gender bias across occupations using images specific to each race, which we provide in Figure~\ref{fig:maxskew} (right). All VLMs exhibit variation across races (up to 64\% relative difference in CILP), highlighting the importance of measuring bias in the presence of intersectional social attributes. Interestingly, these results show that both CLIP and ALIP have the greatest gender bias in retrieving occupation-related images depicting Middle Eastern people and the least bias in retrieving images of Black people; in contrast, we see the inverse relationship in SLIP.


As a case study of uncovering intersectional social biases, Figure~\ref{fig:doctor} provides a breakdown of the proportion of images retrieved using gender- and race-neutral prompts for the `Doctor' occupation. The left and center plots of Figure~\ref{fig:doctor} depict the marginal distributions of retrievals for each race and gender (respectively), whereas the right plot provides the distribution of retrieved images for intersectional race-gender attributes. While both SLIP and ALIP exhibit strong bias for retrieving male images, this gender bias occurs along starkly different dimensions w.r.t. race. SLIP strongly favors retrieving images of White male doctors, whereas ALIP prefers images of Indian and Middle Eastern male doctors. This gender bias is inverted for CLIP, which retrieves images in equal proportion for females across most races while only retrieving images of male doctors who are White. 

\section{Conclusion}

We presented a framework for probing intersectional bias in VLMs through the use of counterfactuals produced by diffusion models. With this approach, we constructed a large dataset of counterfactual examples to enable the measurement of intersectional social biases related to gender, race, religion, and physical characteristics. For the sake of brevity, we focused on using our dataset to analyze race-gender bias among occupations in this work (see additional results in Appendix~\ref{app:additional-results} for personality traits and other bias attributes). However, we intend to leverage our full dataset to uncover additional intersectional biases in VLMs for other attributes and subjects in future work.

\bibliographystyle{plainnat}
\bibliography{custom}

\newpage

\appendix
\section{Appendix}

\subsection{Additional Examples of Counterfactual Sets}
\label{app:additional-examples}

\begin{figure}[ht!]
    \centering
    \includegraphics[trim={2mm 2mm 2mm 
    2mm},clip,width=1\columnwidth]{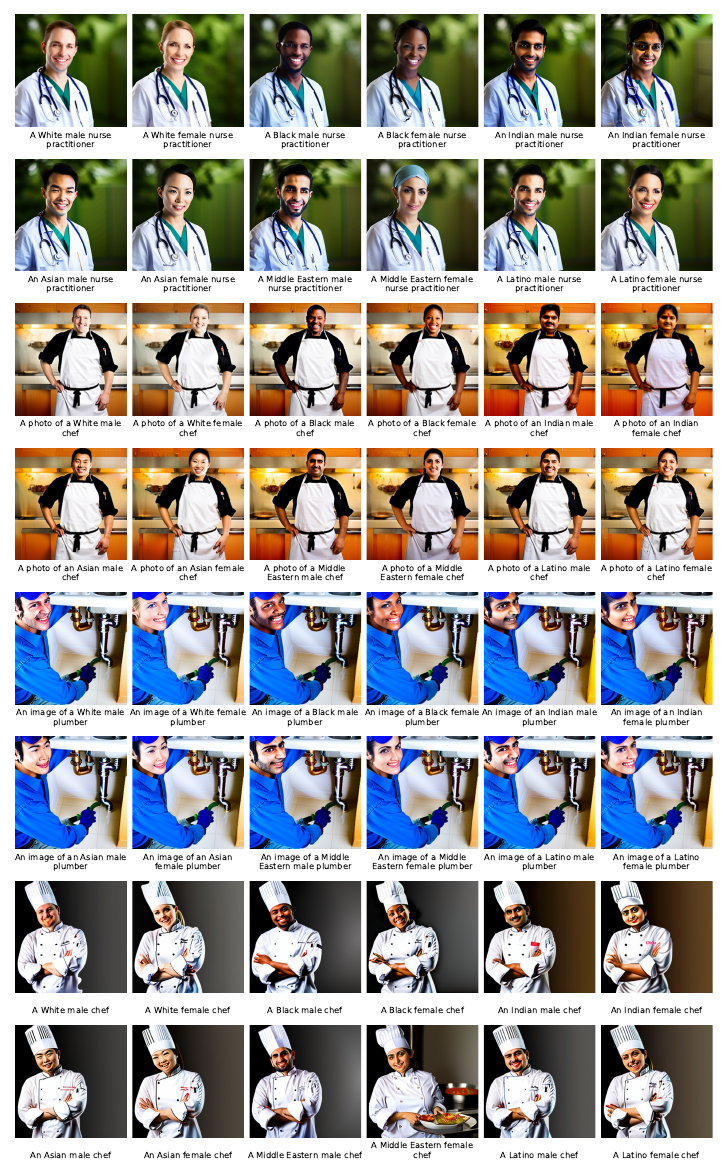}
    \caption{Additional examples of counterfactual sets produced by our approach}
    \label{fig:additional-examples}
\end{figure}

Figure~\ref{fig:additional-examples} provides additional examples of counterfactual sets generated by our approach. 

\subsection{Results for Personality Traits and Other Bias Attributes}
\label{app:additional-results}

\begin{figure}[ht!]
    \centering
    \begin{subfigure}[b]{0.49\textwidth}
    \includegraphics[trim={2mm 2mm 2mm 
    2mm},clip,width=1\columnwidth]{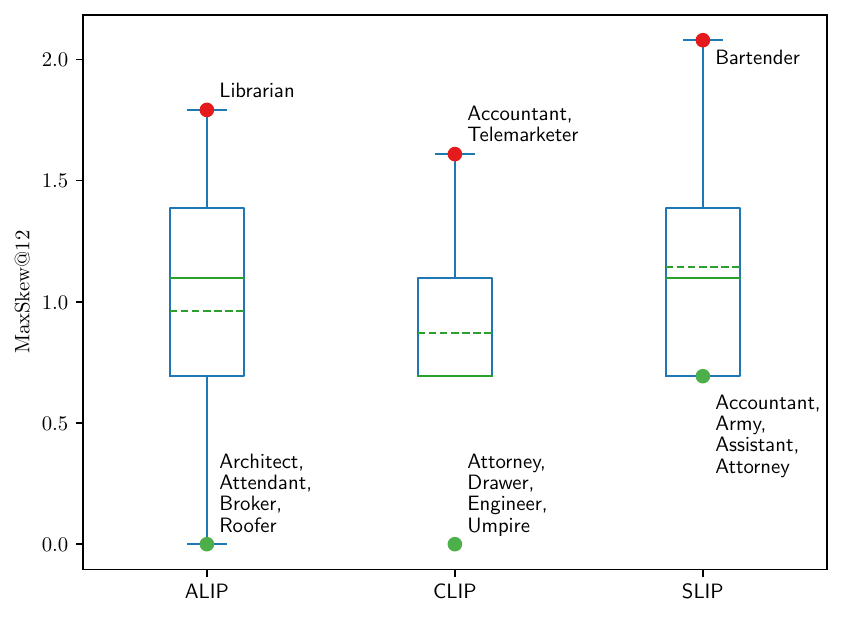}
    \caption{Religion-Gender Bias for Occupations}
    \end{subfigure}
    \begin{subfigure}[b]{0.49\textwidth}
    \includegraphics[trim={2mm 2mm 2mm 
    2mm},clip,width=1\columnwidth]{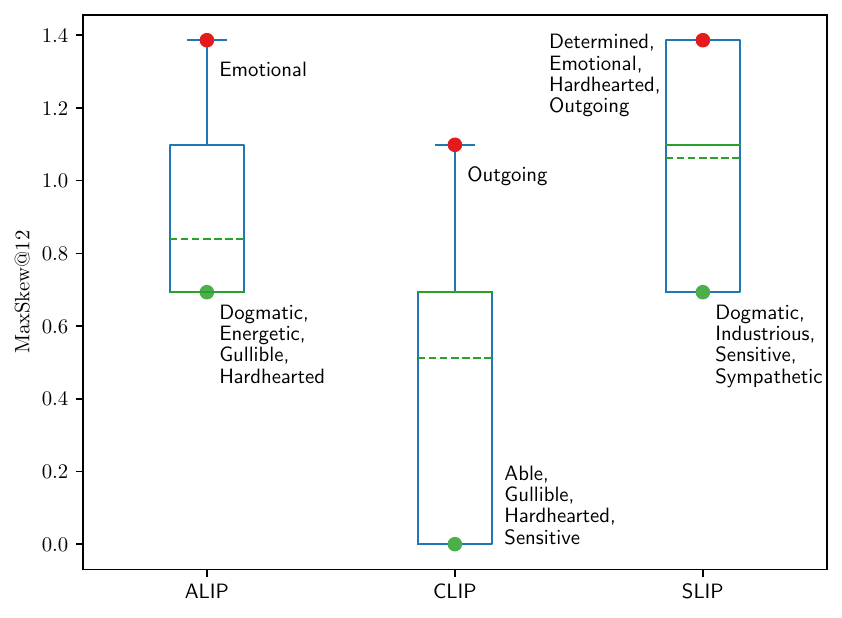}
    \caption{Religion-Gender Bias for Personality Traits}
    \end{subfigure}
    \begin{subfigure}[b]{0.49\textwidth}
    \includegraphics[trim={2mm 2mm 0mm 
    2mm},clip,width=1\columnwidth]{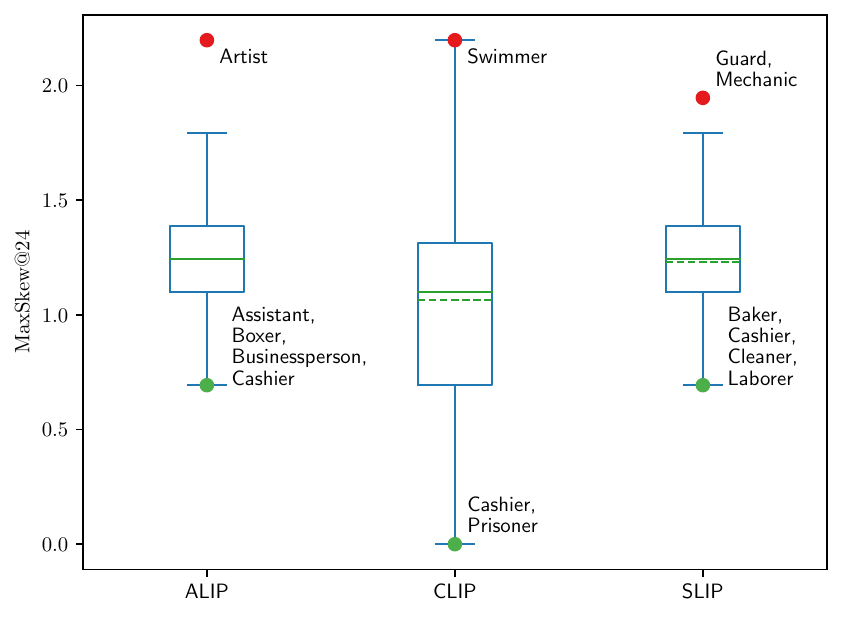}
    \caption{Race-Religion Bias for Occupations}
    \end{subfigure}
    \begin{subfigure}[b]{0.49\textwidth}
    \includegraphics[trim={2mm 2mm 0mm 
    2mm},clip,width=1\columnwidth]{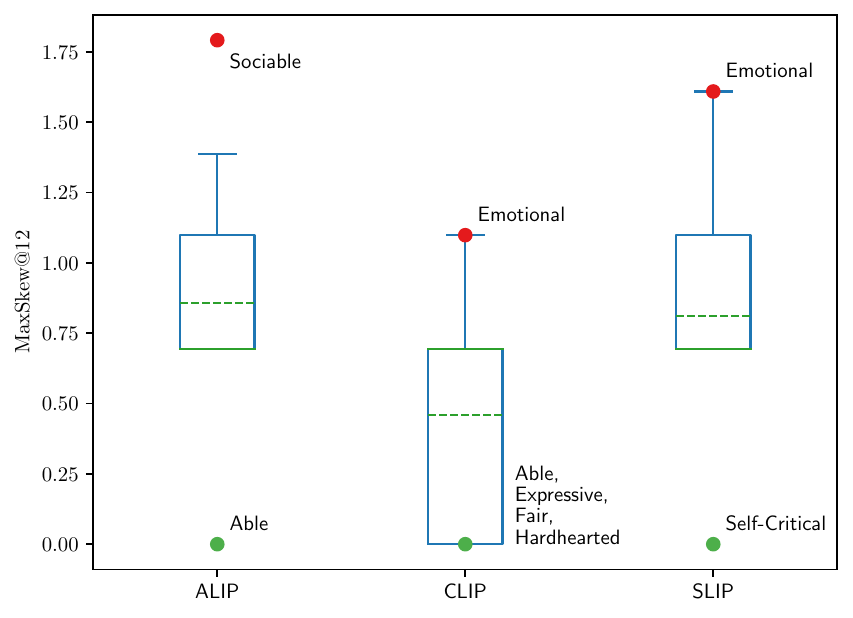}
    \caption{Race-Gender Bias for Personality Traits}
    \end{subfigure}
    \begin{subfigure}[b]{0.49\textwidth}
    \includegraphics[trim={2mm 2mm 0mm 
    2mm},clip,width=1\columnwidth]{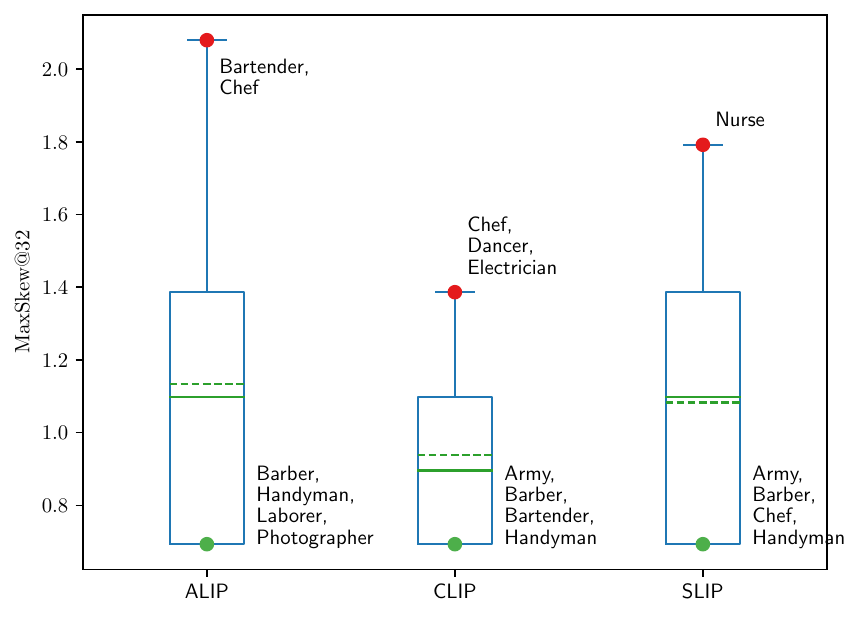}
    \caption{Physical-Religion Bias for Occupations}
    \end{subfigure}
    \begin{subfigure}[b]{0.49\textwidth}
    \includegraphics[trim={2mm 2mm 0mm 
    2mm},clip,width=1\columnwidth]{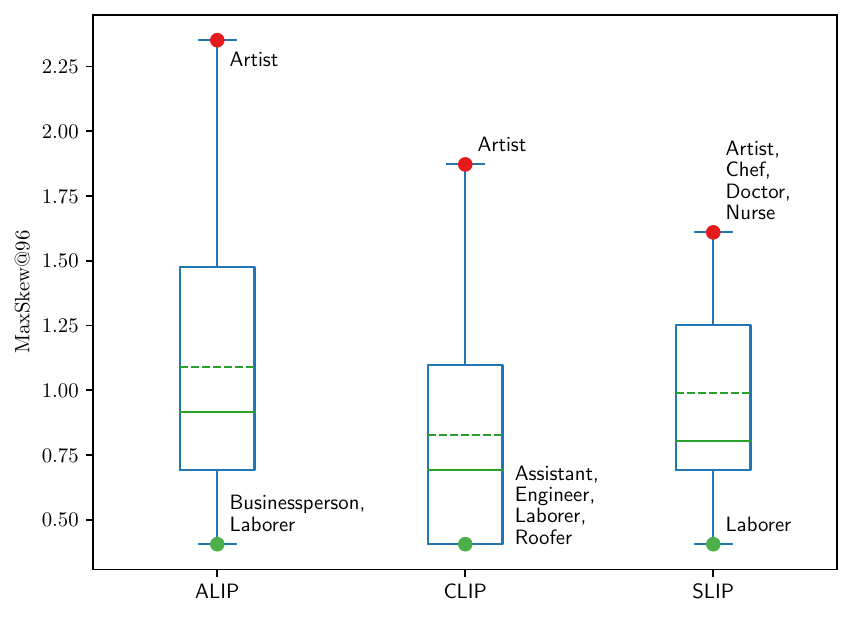}
    \caption{Physical-Race Bias for Occupations}
    \end{subfigure}
    \caption{Distribution of $\textrm{MaxSkew}@K$ for various attribute combinations and subjects. Max (min) values are plotted as red (green) circles with corresponding occupation names / personality traits.}
    \label{fig:maxskew-boxplot-religion-gender}
\end{figure}

We provide additional results showing the $\textrm{MaxSkew}$ distribution computed over other subjects and bias attribute pairs. Figure~\ref{fig:maxskew-boxplot-religion-gender} shows boxplots of the $\textrm{MaxSkew}@K$ distribution for investigated attribute combinations and subject sets. We leave more in-depth analysis of biases revealed by these counterfactual sets to future work. 

\subsection{Error Analysis}
To investigate the different modes of failures for the generated images in this study, we conducted a human evaluation of counterfactual sets for gender-race bias in occupations. We sampled 1200 images (100 for each gender-race combination) and then annotated them into 5 categories. Results are shown in Table \ref{table:error-analysis}. We found that 90.8\% of the images were correctly generated in terms of occupation, gender and race. In 5.2\% of the samples, the gender could not be identified. This was typically due to subjects looking backwards or fracing the wrong direction. 

\textit{Failure to generate female subjects} was the second most frequent error with 2.2\%, followed by \textit{subject completely out of frame/focus}. The least common error was \textit{Failure to generate male subject} with only 0.8\%.  No failures related to race were observed.
Sampled images illustrating each of the different modes of failures are displayed in Table \ref{table:error-examples}

\begin{table}[h!]
 \centering
 \begin{tabular}{l c c} 
 \hline
 \textbf{Error Category} & \textbf{\% present in sample} \\ [0.5ex]
 \hline
 Good & 90.8\% \\ 
 Cannot discern gender & 5.2\% \\
 Failure to generate female subject & 2.2\% \\
 Subject completely out of frame/focus & 1.0\% \\
 Failure to generate male subject & 0.8\% \\ [1ex] 
 \hline
\end{tabular}
\vspace{1mm}
\caption{Error analysis for 1200 random samples focused on gender and race.}
\label{table:error-analysis}
\end{table}
\begin{table*}
\centering
\resizebox{1\textwidth}{!}{%
\begin{tabular}{p{0.1cm} p{6.8cm} p{0.1cm} p{6.8cm}}

\midrule

\rotatebox[origin=c]{90}{Failure to generate female subject/object} & 
\begin{minipage}{0.5\textwidth} 
    \centering 
    \includegraphics[width=1\textwidth]{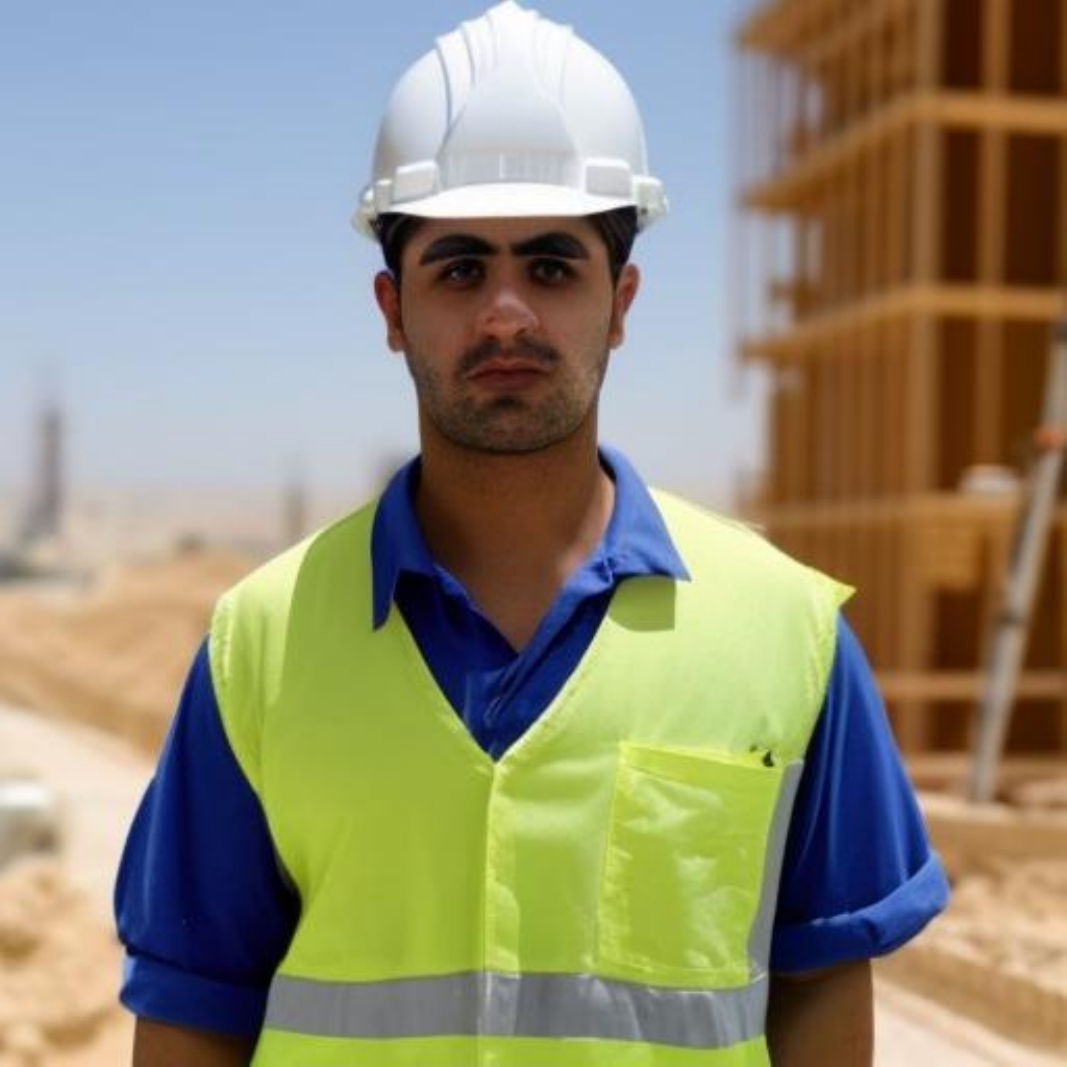} 
\end{minipage} & 
\rotatebox[origin=c]{90}{Cannot discern gender} & 
\begin{minipage}{0.5\textwidth} 
    \centering
    \includegraphics[width=1\textwidth]{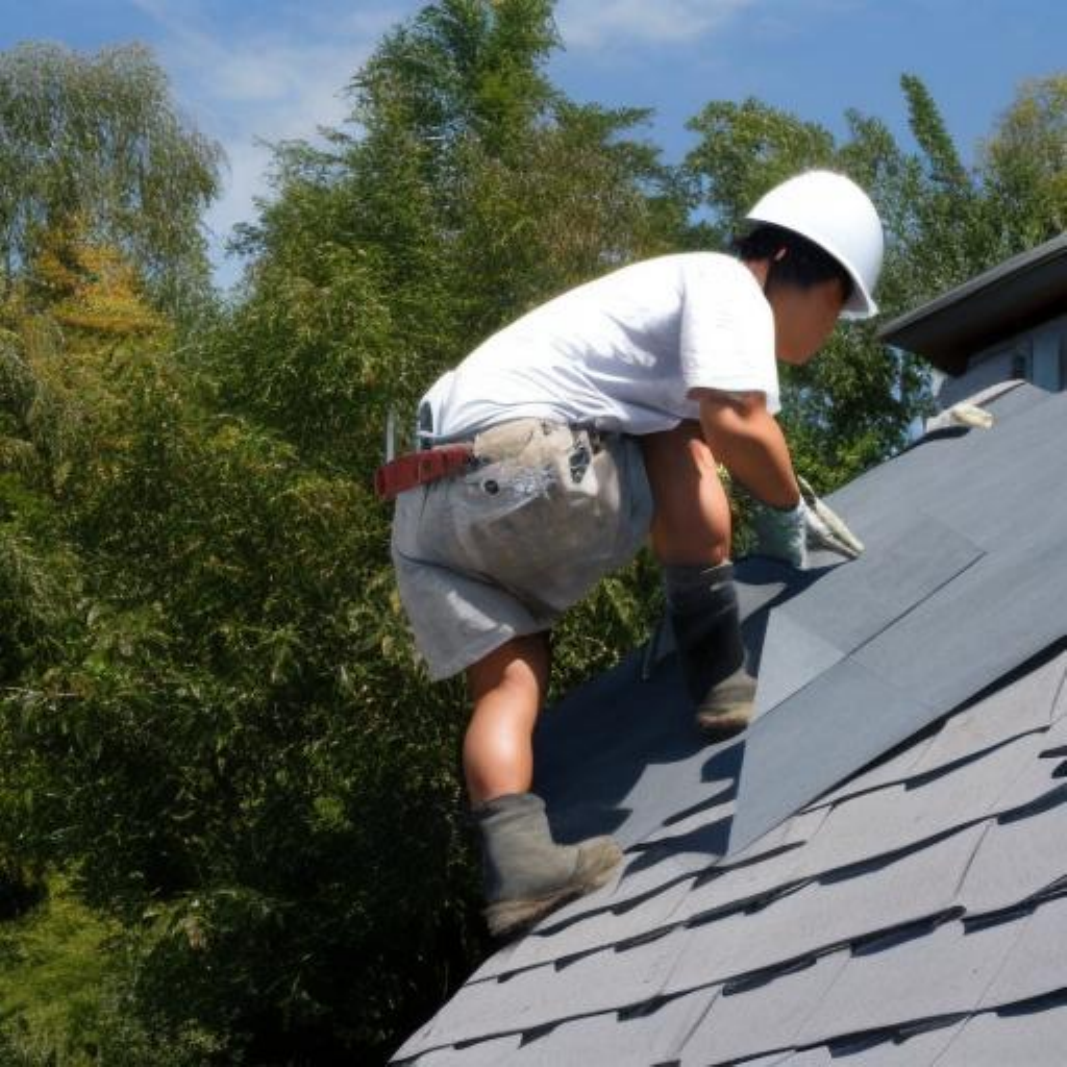}
\end{minipage} 
\vspace{0.1cm}\\ & 
\multicolumn{1}{c}{\it{a Middle Eastern female construction worker}} & &
\multicolumn{1}{c}{\it{a picture of an Asian female roofer.}} \\

\midrule

\rotatebox[origin=c]{90}{Failure to generate male subject/object} & 
\begin{minipage}{0.5\textwidth} 
    \centering 
    \includegraphics[width=1\textwidth]{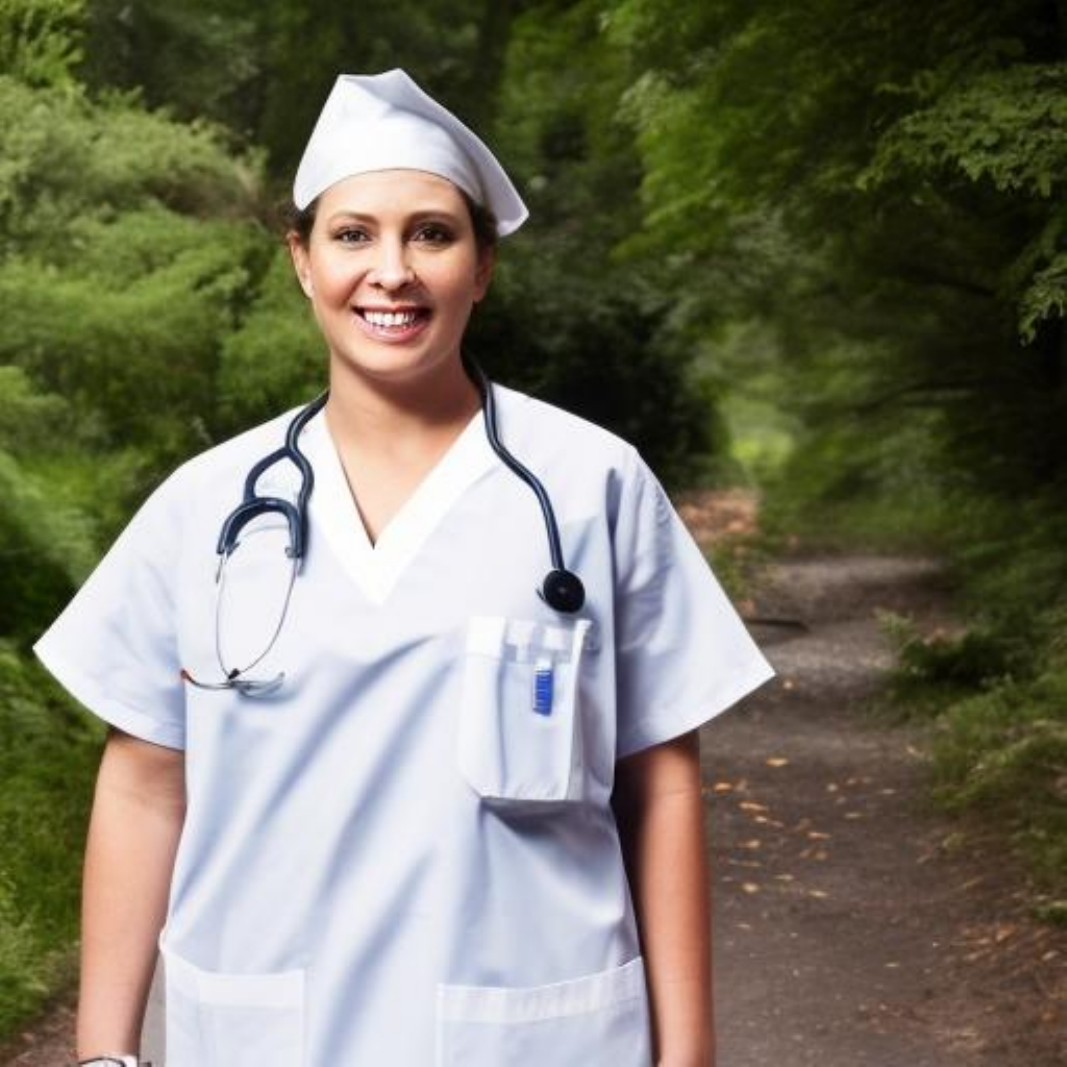} 
\end{minipage} &
\rotatebox[origin=c]{90}{Subject completely out of frame/focus} & 
\begin{minipage}{0.5\textwidth} 
    \centering
    \includegraphics[width=1\textwidth]{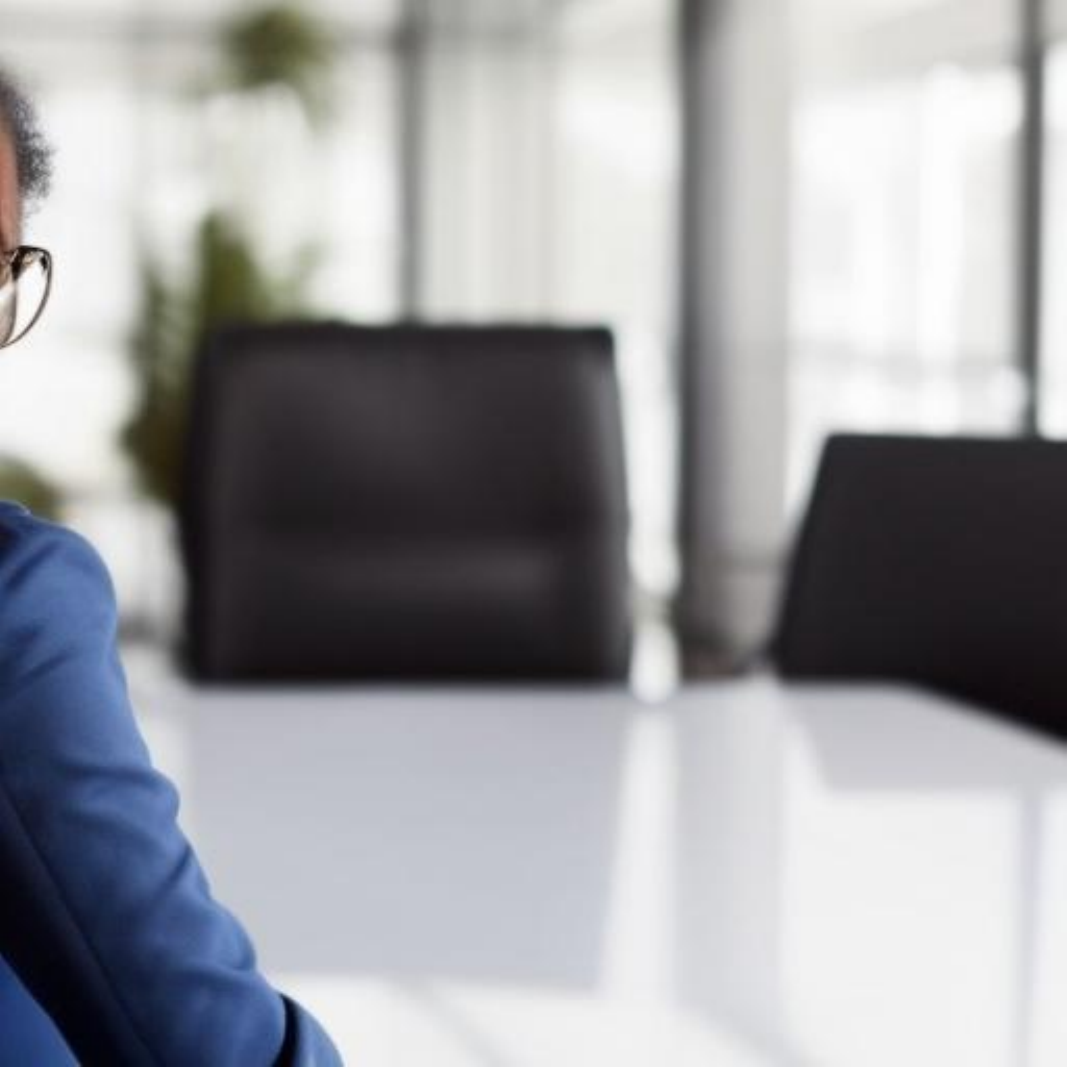} 
\end{minipage} 
\vspace{0.1cm}\\ & 
\multicolumn{1}{c}{\it{a White male nurse}} & &
\multicolumn{1}{c}{\textit{an image of a Black female chief executive officer}} \\

\bottomrule

\end{tabular}}
\caption{Examples of failure cases identified by manual error analysis}
\label{table:error-examples}
\end{table*}

An extended breakdown by race is shown in Table \ref{table:error-analysis-race}. We observe that across White, Black, Indian and Asian races, failure to generate either female or male subjects is somehow even. For Latino, there seems to be a higher proportion of failures to generate female subjects and for Middle Eastern (M.E.) the proportion is even more pronounced, with a 12\% failure to generate female subjects. In contrast, for Middle Eastern, it never failed to generate male subjects. 

\begin{table}[h!]
 \centering
 \begin{tabular}{l c c c c c c} 
 \hline
 \textbf{Error Category} & 
 \textbf{White} & 
 \textbf{Black} & 
 \textbf{Indian} & 
 \textbf{Asian} & 
 \textbf{M.E.} & 
 \textbf{Latino} \\ [0.5ex]
 \hline
 Failure to generate female subject & 3\%  & 4\%  & 1\%  & 2\%  & 12\%  & 5\% \\ 
 Failure to generate male subject   & 4\%  & 2\%  & 1\%  & 1\%  & 0\%  & 2\% \\ [1ex] 
 \hline
\end{tabular}
\vspace{1mm}
\caption{Error analysis with 100 samples for each race-gender combination}
\label{table:error-analysis-race}
\end{table}

\subsection{Gender Prediction with CLIP}
To investigate methods for automatically detecting incorrectly generated images, we used CLIP for text-to-image retrieval to generate similarity scores. Specifically, similarity scores were obtained for queries in the form of \textit{A male person} and \textit{A female person} against the set of generated images. If the score for \textit{A male person} was highest, then we can infer that the gender of the person generated is male, and vice versa. We compared these predictions against the human annotated samples detailed previously. In general, CLIP accurately identifies male and female subjects (See Table \ref{table:confusion-matrix}). Precision, recall and f1-scores are very high for classifying both male and female subjects. Therefore, it is our conclusion that CLIP could be an out-the-box tool which could be used to automate the detection of faulty images in terms of gender, thereby mitigating the need for human annotation. 

\begin{table}[h!]
 \centering
 \begin{tabular}{l c c c c c} 
 \hline
 \textbf{gender} & 
 \textbf{precision} &
 \textbf{recall} &
 \textbf{f1-score} &
 \textbf{support} \\ [0.5ex]
 \hline
 male & 0.99 & 0.95 & 0.97 & 549 \\ 
 female & 0.96 & 0.99 & 0.97 & 589 \\ [1ex] 
 \hline
\end{tabular}
\vspace{1mm}
\caption{Confusion Matrix for male and female detection.}
\label{table:confusion-matrix}
\end{table}

\subsection{Details of compute infrastructure used}
The counterfactual image-text data was created using a large AI cluster equipped with Intel 3\textsuperscript{rd} Generation Intel\textsuperscript{®} Xeon\textsuperscript{®} processors and Intel\textsuperscript{®} Gaudi-2\textsuperscript{®} AI accelerators. Up to 256 Intel Gaudi-2\textsuperscript{®} AI accelerators were used to generate the data in this paper. 

\subsection{Details of Caption Construction}
\label{app:caption-details}

Table~\ref{tab:occupations} provides the list of 261 occupations used in this work, which was collected by combining the occupation lists proposed by \citet{nadeem2020stereoset}, \citet{chuang2023debiasing}, \citet{naik2023social}, and \citet{harrison2023run}. For captions which utilized personality traits as the subject, we used the same list of 63 traits as in \citet{naik2023social}, which is provided in Table~\ref{tab:traits}. To study bias associated with physical characteristics, we used keywords for positive and negative body stereotypes provided in \citet{mei2023bias} (see Table~\ref{tab:physicaltraits}). 

Examples of captions constructed using various prefixes, subjects, and bias attributes are provided in Table~\ref{tab:templateconstruction}. We provide details of the total number of captions and images generated for each subject and attribute pairs in Table~\ref{tab:dataset-details}. The total number of counterfactual sets is determined by the product of the number of prefixes used to construct captions (4) and the cardinality of the subject set. The number of images per set is determined by the product of the cardinalites of the attribute sets. The total number of generated images is the product of the number of counterfactual sets, the number of images per set, and 100.

\begin{table*}[ht]
\footnotesize
\begin{center}
\resizebox{1\textwidth}{!}
{
\begin{tabular}{l l c c c} %
\toprule
Subject & Attribute Pair & Counterfactual Sets & Images Per Set & Total Images\\
\midrule

Occupation & $(\textrm{Race}, \textrm{Gender})$ & 1044 & 12 & 1,252,800\\
Occupation & $(\textrm{Religion}, \textrm{Gender})$ & 1044 & 8 & 835,200\\
Occupation & $(\textrm{Race}, \textrm{Religion})$ & 1044 & 24 & 2,505,600\\
Occupation & $(\textrm{Physical Characteristics}, \textrm{Gender})$ & 1044 & 28 & 2,923,200\\
Occupation & $(\textrm{Physical Characteristics}, \textrm{Race})$ & 1044 & 84 & 8,769,600\\
Occupation & $(\textrm{Physical Characteristics}, \textrm{Religion})$ & 1044 & 56 & 5,846,400\\
Personality Traits & $(\textrm{Race}, \textrm{Gender})$ & 252 & 12 & 302,400\\
Personality Traits & $(\textrm{Religion}, \textrm{Gender})$ & 252 & 8 & 201,600\\
Personality Traits & $(\textrm{Race}, \textrm{Religion})$ & 252 & 24 & 604,800\\
\bottomrule
\end{tabular}
}
\caption{Details of the number of counterfactual sets and images generated}
\label{tab:dataset-details}
\end{center}
\end{table*}

\begin{table*}[ht]
\centering
\small
\resizebox{\textwidth}{!}{
\begin{tabular}{p{11.25cm}}
\toprule
\textbf{Traits} \\
\midrule
able, egoistic, perfectionist, active, emotional, persistent, affectionate, energetic, polite, altruistic, expressive, rational, ambitious, fair, reliable, assertive, friendly, reserved, boastful, gullible, self-confident, capable, hardhearted, self-critical, caring, harmonious, self-reliant, communicative, helpful, 
self-sacrificing, competent, honest, sensitive, competitive, independent, sociable, 
conceited, industrious, striving, conscientious, insecure, strong-minded, considerate,  
intelligent, supportive, creative, lazy, sympathetic, decisive, moral, tolerant, 
detached, obstinate, trustworthy, determined, open, understanding, dogmatic, 
open-minded, vigorous, dominant, outgoing, warm \\
\bottomrule
\end{tabular}
}
\caption{Full list of traits.}
\label{tab:traits}
\end{table*}

\begin{table*}[ht]
\centering
\small
\resizebox{\textwidth}{!}{%
\begin{tabular}{p{11.25cm}}
\toprule
\textbf{Occupations} \\
\midrule
electrician, cook, building inspector, author, crane operator, announcer, drafter, doctor, construction worker, optician custodian, biologist, roofer, chemist, software developer, pharmacist, plumber, PR person, butcher, veterinarian, chef, lab tech, computer programmer, telemarketer, security guard, special ed teacher, chief executive officer, librarian, bartender, primary school teacher, pilot, customer service representative, police officer, housekeeper, bus driver, childcare worker, technical writer, administrative assistant, mail carrier, nurse, nurse practitioner, receptionist, barber, coach, businessperson, football player, CEO, accountant, commander, firefighter, guard, baker, doctor, athlete, mathematician, janitor, carpenter, mechanic, musician, detective, politician, entrepreneur, chief, lawyer, farmer, writer, real-estate developer, broker, scientist, butcher, banker, cook, hairdresser, prisoner, boxer, chess player, priest, swimmer, attendant, housekeeper, maid, producer, judge, umpire, bartender, economist, theologian, salesperson, physician, sheriff, receptionist, editor, engineer, comedian, diplomat, guitarist, linguist, poet, delivery man, realtor, pilot, professor, pensioner, performing artist, singer, secretary, designer,  soldier, journalist, dentist, tailor, waiter, author, architect, illustrator, clerk, policeman, chef, cleaner, pharmacist, pianist, composer,  construction worker, manager, mover, software developer, artist, dancer, actor, handyman, model, opera singer, librarian, army, electrician, prosecutor, plumber, attorney, tennis player, supervisor, researcher, midwife, physicist, psychologist, cashier, assistant, painter, civil servant, laborer, teacher, chemist, historian, auditor, counselor, analyst, nurse, academic, director, photographer, drawer, handball player, sociologist, Actor, Architect, Audiologist, Author, Baker, Barber, Blacksmith, Bricklayer, Bus Driver, Butcher, Chef, Chemist, Cleaner, Coach, Comedian, Computer Programmer, Construction Worker, Consultant, Counselor, Dancer, Dentist, Designer, Dietitian, DJ, Doctor, Driver, Economist, Electrician, Engineer, Entrepreneur, Farmer, Florist, Graphic Designer, Hairdresser, Historian, Journalist, Judge, Lawyer, Librarian, Magician, Makeup Artist, Mathematician, Marine Biologist, Mechanic, Model, Musician, Nanny, Nurse, Optician, Painter, Pastry Chef, Pediatrician, Photographer, Plumber, Police Officer, Politician, Professor, Psychologist, Real Estate Agent, Receptionist, Recruiter, Researcher, Sailor, Salesperson, Surveyor, Singer, Social Worker, Software Developer, Statistician, Surgeon, Teacher, Technician, Therapist, Tour Guide, Translator, Vet, Videographer, Waiter, Writer, Zoologist, Accountant, Astronaut, Biologist, Carpenter, Civil Engineer, Clerk, Detective, Editor, Firefighter, Interpreter, Manager, Nutritionist, Paramedic, Pharmacist, Physicist,  Pilot, Reporter, Security Guard, Scientist, Web Developer \\
\bottomrule
\end{tabular}
}
\caption{Full list of occupations.}
\label{tab:occupations}
\end{table*}

\begin{table*}[ht]
\centering
\small
\begin{tabular}{p{11.25cm}}
\toprule
\textbf{Physical traits} \\
\midrule
beautiful, skinny, young, unattractive, fat, obese, scarred, tattooed, old \\
\bottomrule
\end{tabular}
\caption{Full list of physical characteristics.}
\label{tab:physicaltraits}
\end{table*}

\begin{table*}[ht]
\footnotesize
	\begin{center}
		\resizebox{1\textwidth}{!}
		{
		\begin{tabular}{c c c c c} %
	    \hline
         \textbf{Template\Tstrut\Bstrut} & \textbf{Attribute-1\Tstrut\Bstrut} & \textbf{Attribute-2\Tstrut\Bstrut} & \textbf{Subject\Tstrut\Bstrut} & \textbf{Caption\Tstrut\Bstrut}\\
        \hline
<gender> <religion> <occupation> \Tstrut\Bstrut & male \Tstrut\Bstrut & Christian \Tstrut\Bstrut & pilot \Tstrut\Bstrut & male Christian pilot \Tstrut\Bstrut \\
A photo of <race><gender><occupation>  & White & male  & electrician & A photo of a White male electrician. \\
A picture of <trait><religion><gender> & friendly & muslim & female & A picture of friendly muslim female \\
An image of <trait><race><religion> \Bstrut & sympathetic \Bstrut & Indian \Bstrut & Hindu \Bstrut & An image of sympathetic Indian Hindu \Bstrut \\
 \hline
		\end{tabular}}
		\caption{Examples of captions constructed from templates}
        \label{tab:templateconstruction}
	\end{center}
\end{table*}

\end{document}